\documentstyle[fleqn,leqno]{article}

\catcode`@=11

\long\def\xavm[#1]#2{\left[ \begin{array}{l}#2 %
                \end{array} \right]_{#1} \vspace{0.5ex}}

\def\avm{\@ifnextchar[{\xavm}{\xavm[{}]}}

\def\xvar[#1]#2{\mbox{\rm #2}_{#1}}
\def\var{\@ifnextchar[{\xvar}{\xvar[{}]}}

\def\xatt[#1]#2{\mbox{\it #2}_{#1}}
\def\att{\@ifnextchar[{\xatt}{\xatt[{}]}}

\def\xcon[#1]#2{\mbox{\rm #2}_{#1}}
\def\con{\@ifnextchar[{\xcon}{\xcon[{}]}}

  \newlength\titlebox 
 \twocolumn 

\def\addcontentsline#1#2#3{}

\def\maketitle{\par
 \begingroup
   \def\thefootnote{\fnsymbol{footnote}}
   \def\@makefnmark{\hbox to 0pt{$^{\@thefnmark}$\hss}}
   \twocolumn[\@maketitle] \@thanks
 \endgroup
 \setcounter{footnote}{0}
 \let\maketitle\relax \let\@maketitle\relax
 \gdef\@thanks{}\gdef\@author{}\gdef\@title{}\let\thanks\relax}
\def\@maketitle{\vbox to \titlebox{\hsize\textwidth
 \linewidth\hsize \vskip 0.625in minus 0.125in \centering
 {\huge\bf \@title \par} \vskip 0.2in plus 1fil minus 0.1in
 {\def\and{\unskip\enspace{\rm and}\enspace}%
  \def\And{\end{tabular}\hss \egroup \hskip 1in plus 2fil
           \hbox to 0pt\bgroup\hss \begin{tabular}[t]{c}\Large\bf}%
  \def\AND{\end{tabular}\hss\egroup \hfil\hfil\egroup
	  \vskip 0.25in plus 1fil minus 0.125in
	   \hbox to \linewidth\bgroup\Large \hfil\hfil
 	     \hbox to 0pt\bgroup\hss \begin{tabular}[t]{c}\Large\bf}
  \hbox to \linewidth\bgroup\Large \hfil\hfil
    \hbox to 0pt\bgroup\hss \begin{tabular}[t]{c}\Large\bf\@author
			    \end{tabular}\hss\egroup
    \hfil\hfil\egroup}
  \vskip 0.3in plus 2fil minus 0.1in
}}

\def\section{\@startsection {section}{1}{\z@}{-2.0ex plus
    -0.5ex minus -.2ex}{3pt plus 2pt minus 1pt}{\Large\bf\centering}}
\def\subsection{\@startsection{subsection}{2}{\z@}{-2.0ex plus
    -0.5ex minus -.2ex}{3pt plus 2pt minus 1pt}{\large\bf\raggedright}}
\def\subsubsection{\@startsection{subparagraph}{3}{\z@}{-6pt plus
   2pt minus 1pt}{-1em}{\normalsize\bf}}

\skip\footins 9pt plus 4pt minus 2pt
\def\footnoterule{\kern-3pt \hrule width 5pc \kern 2.6pt }
\labelwidth\leftmargini\advance\labelwidth-\labelsep \labelsep 5pt

\def\@listi{\leftmargin\leftmargini
	\labelwidth\leftmargini\advance\labelwidth-\labelsep}
\def\@listii{\leftmargin\leftmarginii
   \labelwidth\leftmarginii\advance\labelwidth-\labelsep
   \topsep 2pt plus 1pt minus 0.5pt
   \parsep 1pt plus 0.5pt minus 0.5pt
   \itemsep \parsep}
\def\@listiii{\leftmargin\leftmarginiii
    \labelwidth\leftmarginiii\advance\labelwidth-\labelsep
    \topsep 1pt plus 0.5pt minus 0.5pt
    \parsep \z@ \partopsep 0.5pt plus 0pt minus 0.5pt
    \itemsep \topsep}
\def\@listiv{\leftmargin\leftmarginiv
     \labelwidth\leftmarginiv\advance\labelwidth-\labelsep}
\def\@listv{\leftmargin\leftmarginv
     \labelwidth\leftmarginv\advance\labelwidth-\labelsep}
\def\@listvi{\leftmargin\leftmarginvi
     \labelwidth\leftmarginvi\advance\labelwidth-\labelsep}

\belowdisplayskip \abovedisplayskip
\def\small{\@setsize\small{10pt}\ixpt\@ixpt}
\def\footnotesize{\@setsize\footnotesize{10pt}\ixpt\@ixpt}
\def\scriptsize{\@setsize\scriptsize{8pt}\viipt\@viipt}
\def\tiny{\@setsize\tiny{7pt}\vipt\@vipt}
\def\Large{\@setsize\Large{14pt}\xiipt\@xiipt}
\def\LARGE{\@setsize\LARGE{16pt}\xivpt\@xivpt}
\def\huge{\@setsize\huge{20pt}\xviipt\@xviipt}
\def\Huge{\@setsize\Huge{23pt}\xxpt\@xxpt}

\def\@citex[#1]#2{\if@filesw\immediate\write\@auxout{\string\citation{#2}}\fi
  \def\@citea{}\@cite{\@for\@citeb:=#2\do
    {\@citea\def\@citea{;\penalty\@m\ }\@ifundefined
       {b@\@citeb}{{\bf ?}\@warning
       {Citation `\@citeb' on page \thepage \space undefined}}%
{\csname b@\@citeb\endcsname}}}{#1}}
\let\@internalcite\cite
\def\cite{\def\citename##1{##1, }\@internalcite}
\def\shortcite{\def\citename##1{}\@internalcite}
\def\newcite{\leavevmode\def\citename##1{{##1} (}\@internalciteb}

\def\@citexb[#1]#2{\if@filesw\immediate\write\@auxout{\string\citation{#2}}\fi
  \def\@citea{}\@newcite{\@for\@citeb:=#2\do
    {\@citea\def\@citea{;\penalty\@m\ }\@ifundefined
       {b@\@citeb}{{\bf ?}\@warning
       {Citation `\@citeb' on page \thepage \space undefined}}%
\hbox{\csname b@\@citeb\endcsname}}}{#1}}
\def\@internalciteb{\@ifnextchar [{\@tempswatrue\@citexb}
					{\@tempswafalse\@citexb[]}}

\def\@newcite#1#2{{#1\if@tempswa, #2\fi)}}

\def\@biblabel#1{\def\citename##1{##1}[#1]\hfill}

\def\@cite#1#2{({#1\if@tempswa , #2\fi})}

\def\thebibliography#1{\vskip\parskip%
\vskip\baselineskip%
\def\baselinestretch{1}%
\vskip-\parskip%
\vskip-\baselineskip%
\section*{References\@mkboth
 {References}{References}}\list
 {}{\setlength{\labelwidth}{0pt}\setlength{\leftmargin}{\parindent}
 \setlength{\itemindent}{-\parindent}}
 \def\newblock{\hskip .11em plus .33em minus -.07em}
 \sloppy\clubpenalty4000\widowpenalty4000
 \sfcode`\.=1000\relax}

\def\thesourcebibliography#1{\vskip\parskip%
\vskip\baselineskip%
\def\baselinestretch{1}%
\ifx\@currsize\normalsize\@normalsize\else\@currsize\fi%
\vskip-\parskip%
\vskip-\baselineskip%
\section*{Sources of Attested Examples\@mkboth
 {Sources of Attested Examples}{Sources of Attested Examples}}\list
 {}{\setlength{\labelwidth}{0pt}\setlength{\leftmargin}{\parindent}
 \setlength{\itemindent}{-\parindent}}
 \def\newblock{\hskip .11em plus .33em minus -.07em}
 \sloppy\clubpenalty4000\widowpenalty4000
 \sfcode`\.=1000\relax}

\def\@lbibitem[#1]#2{\item[]\if@filesw
      { \def\protect##1{\string ##1\space}\immediate
        \write\@auxout{\string\bibcite{#2}{#1}}\fi\ignorespaces}}

\def\@bibitem#1{\item\if@filesw \immediate\write\@auxout
       {\string\bibcite{#1}{\the\c@enumi}}\fi\ignorespaces}

\title{Constraint-Based Categorial Grammar}

\author{Gosse Bouma and Gertjan van Noord\\
Alfa-informatica and \\
Behavorial and Cognitive Neurosciences,
\\Rijksuniversiteit Groningen\\
\{gosse,vannoord\}@let.rug.nl}

\begin{document}

\maketitle

\abstract{We propose a generalization of Categorial Grammar in which lexical
categories are defined by means of recursive constraints.  In particular, the
introduction of relational constraints allows one to capture the effects of
(recursive)
lexical rules in a computationally attractive manner.  We illustrate the
linguistic merits of the new approach by showing how it accounts for the syntax
of
Dutch cross-serial dependencies and the position and scope of adjuncts in such
constructions.  Delayed evaluation is used to process grammars containing
recursive constraints.}

\section{Introduction}

Combinations of Categorial Grammar ({\sc cg}) and unification naturally lead to
the introduction of polymorphic categories.  Thus, Karttunen
\shortcite{karttunen:89} categorizes {\sc np}'s as {\sc x}$/${\sc x}, where
{\sc
x} is a verbal category, Zeevat {\em et al.}  \shortcite{ucg} assign the
category {\sc x}$/$({\sc np}$\backslash${\sc x}) to {\sc np}'s, and Emms
\shortcite{emms:93} extends the Lambek-calculus with polymorphic categories to
account for coordination, quantifier scope, and extraction.

The role of polymorphism has been restricted, however, by the fact that in
previous work categories were defined as feature structures using the simple,
non-recursive, constraints familiar from feature description languages such as
{\sc patr}.  Relational constraints can be used to define a range of
polymorphic
categories that are beyond the expressive capabilities of previous approaches.

In particular, the introduction of relational constraints captures the effects
of (recursive) lexical rules in a computationally attractive manner.  The
addition of such rules makes it feasible to consider truly `lexicalist'
grammars, in which a powerful lexical component is accompanied by a highly
restricted syntactic component, consisting of application only.

\section{Recursive Constraints}

In {\sc cg}, many grammatical concepts can only be defined recursively.  Dowty
\shortcite{dowty:82} defines grammatical functions such as {\em subject} and
{\em object} as being the ultimate and penultimate `argument-in' of a verbal
category.  Hoeksema \shortcite{hoeksema-diss} defines verbs as exocentric
categories reducible to {\sc s}.  Lexical rules frequently refer to such
concepts.  For instance, a categorial lexical rule of passive applies to {\em
verbs} selecting an {\em object} and must remove the {\em subject}.

In standard unification-based formalisms, these concepts and the rules
referring
to such concepts cannot be expressed directly.

\subsection{Subject-verb agreement}

Consider a categorial treatment of subject-verb agreement with
intransitive {\sc (~np[nom]$\backslash$s~)} and transitive (~{\sc
(np[nom]$\backslash$s$)/$np[acc]}~) verbs defined as follows:

\begin{equation}
\begin{array}[t]{l}
\att{lex}(\con{walks},\con{X}) \con{ :-} \\
\hspace{1cm} \att{iv}(\con{X}).\\
\att{lex}(\con{kisses},\con{X}) \con{ :-} \\
\hspace{1cm} \att{tv}(\con{X}). \\ \\
\att{iv}(\avm{	\att{val } \avm{\att{cat } \con{s}}\\
				\att{dir } `\backslash\mbox{'}\\
				\att{arg } \avm{\att{cat } \con{np}\\
						\att{case } \con{nom}}}).\\ \\
\att{tv}(
\avm{	\att{val } \avm{ 	\att{val } \avm{\att{cat } \con{s}}\\
				\att{dir } `\backslash\mbox{'}\\
				\att{arg } \avm{\att{cat } \con{np}\\
						\att{case } \con{nom}}}\\
	\att{dir } `/\mbox{'}\\
	\att{arg } \avm{\att{cat } \con{np}\\
			 \att{case } \con{acc}}}).
\end{array}\end{equation}

Subject-verb agreement can be incorporated easily if one reduces agreement to a
form of subcategorization.  If, however, one wishes to distinguish these two
pieces of information (to avoid a proliferation of subcategorization types or
for morphological reasons, for instance), it is not obvious how this could be
done without recursive constraints.  For intransitive verbs one needs the
constraint that $\langle${\em arg agr}$\rangle$ = {\em Agr} (where {\em Agr} is
some agreement value), for transitive verbs that $\langle${\em val arg
agr}$\rangle$ = {\em Agr}, and for ditransitive verbs that $\langle${\em val
val
arg agr}$\rangle$ = {\em Agr}.  The generalization is captured using the
recursive constraint {\em sv\_agreement} (\ref{sva2}).  In (\ref{sva2}) and
below, we use definite clauses to define lexical entries and constraints.  Note
that lexical entries relate words to feature structures that are defined
indirectly as a combination of simple constraints (evaluated by means of
unification) and recursive constraints.\footnote{We use $X/Y$ and $Y\backslash
X$ as shorthand for $\avm{\att{val } \var{X} \\ \att{dir } `/\mbox{'} \\
\att{arg } \var{Y}}$ and $\avm{\att{val } \var{X} \\ \att{dir } `\backslash
\mbox{'} \\ \att{arg } \var{Y}}$, respectively and S, NP, and Adj as `typed
variables' of type $\avm{\att{cat } \con{s}}$, $\avm{\att{cat } \con{np}}$, and
$\avm{\att{cat } \con{adj}}$, respectively.  }

\begin{equation}\label{sva2}
\begin{array}[t]{l}
\att{lex}(\con{walks},\var{X}) \con{ :-}\\
\hspace{1cm} \att{iv}(\var{X}),\\
\hspace{1cm} \att{sv\_agreement}(\con{sg3},\var{X}).\\
\att{lex}(\con{kisses},\var{X}) \con{ :-}\\
\hspace{1cm} \att{tv}(\var{X}),\\
\hspace{1cm} \att{sv\_agreement}(\con{sg3},\var{X}). \\ \\
\att{sv\_agreement}(\con{Agr},
	\avm{\att{cat } \con{np}\\ \att{agr }  \var{Agr}}\backslash \con{S}).\\
\att{sv\_agreement}(\con{Agr},\var{Y}\backslash\var{X}) \con{ :-}\\
\hspace{1cm}	\att{sv\_agreement}(\con{Agr},\var{X}).
\end{array}
\end{equation}

Relational constraints can also be used to capture the effect of lexical rules.
In a lexicalist theory such as {\sc cg}, in which syntactic rules are
considered
to be universally valid scheme's of functor-argument combination, lexical rules
are an essential tool for capturing language-specific generalizations.  As
Carpenter \shortcite{carpenter-lexical} observes, some of the rules that have
been proposed must be able to operate recursively.  Predicative formation in
English, for instance, uses a lexical rule turning a category reducible to {\sc
vp} into a category reducing to a {\sc vp}-modifier {\sc (vp$\backslash$vp)}.
As a {\sc vp}-modifier is reducible to {\sc vp}, the rule can (and sometimes
must) be applied recursively.

\subsection{Adjuncts as arguments}
\label{adjuncts}

Miller \shortcite{miller-diss} proposes a lexical rule for French nouns which
adds an (modifying) adjective to the list of arguments that the noun
subcategorizes for.  Since a noun can be modified by any number of adjectives,
the rule must be optional as well as recursive.  The advantages of using a
lexical rule in this case is that it simplifies accounting for agreement
between
nouns and adjectives and that it enables an account of word order constraints
between arguments and modifiers of a noun in terms of {\em obliqueness}.

The idea that modifiers are introduced by means of a lexical rule can be
extended to verbs.  That is, adjuncts could be introduced by means of a
recursive rule that optionally adds these elements to verbal categories.  Such
a
rule would be an alternative for the standard categorial analysis of adjuncts
as
(endocentric) functors.  There is reason to consider this alternative.

In Dutch, for instance, the position of verb modifiers is not fixed.  Adjuncts
can in principle occur anywhere to the left of the verb:\footnote{As we want to
abstract away from the effects of `verb-second', we present only examples of
subordinate clauses.}

\begin{equation}
\label{adj-ex}
\begin{array}[t]{ll}
a. &
\begin{array}[t]{@{\hspace{0.0cm}}llllll}
\mbox{dat} & \mbox{Johan} & \mbox{opzettelijk} & \mbox{een ongeluk} \\
\mbox{that} & \mbox{J.} & \mbox{deliberately} & \mbox{an accident}
\end{array} \\
& \mbox{veroorzaakt}\\
& \mbox{causes} \\
& \mbox{\em that J. deliberately causes an accident}\\
b. &
\begin{array}[t]{@{\hspace{0.0cm}}llllll}
\mbox{dat} & \mbox{Johan} & \mbox{Marie} & \mbox{opzettelijk}\\
\mbox{that} & \mbox{J.} & \mbox{M.} & \mbox{deliberately}
\end{array} \\
& \begin{array}[t]{@{\hspace{0.0cm}}ll}
  \mbox{geen cadeau} & \mbox{geeft} \\
  \mbox{no present} & \mbox{gives}
  \end{array} \\
& \mbox{\em that J. deliberately gave M. no present}
\end{array}
\end{equation}

There are several ways to account for this fact.  One can assign multiple
categories to adjuncts or one can assign a polymorphic category {\sc x$/$x} to
adjuncts, with {\sc x} restricted to `verbal projections' \cite{bouma:88}.

Alternatively, one can assume that adjuncts are not functors, but arguments of
the verb.  Since adjuncts are optional, can be iterated, and can occur in
several positions, this implies that verbs must be polymorphic.  The constraint
{\em add\_adjuncts} has this effect, as it optionally adds one or more adjuncts
as arguments to the `initial' category of a verb:

\begin{equation}
\label{add-adj}
\begin{array}[t]{l}
\att{lex}(\con{veroorzaken},\var{X}) \con{:-}
\\
\hspace{1cm}
\att{add\_adjuncts}(\var{X},
		\con{NP}\backslash(\con{NP}\backslash\con{S})). \\
\att{lex}(\con{geven},\var{X}) \con{ :-} \\
\hspace{1cm}
\att{add\_adjuncts}(\var{X},
	\con{NP}\backslash(\con{NP}\backslash(\con{NP}\backslash\con{S}))).
\\ \\
\att{add\_adjuncts}(\con{S},\con{S}). \\
\att{add\_adjuncts}(\con{Adj}\backslash \var{X}, \var{Y}) \con{ :-} \\
\hspace{1cm} \att{add\_adjuncts}(\var{X}, \var{Y}). \\
\att{add\_adjuncts}(\avm{	\att{val } \var{X} \\
				\att{dir } \var{D} \\
				\att{arg } \var{A}},
		\avm{	\att{val } \var{Y} \\
				\att{dir } \var{D} \\
				\att{arg } \var{A}}) \con{ :-} \\
\hspace{1cm} \att{add\_adjuncts}(\var{X}, \var{Y}). \\
\end{array}
\end{equation}

This constraint captures the effect of applying the following (schematic)
lexical rule recursively:

\begin{equation}
\begin{array}[t]{c}
X_1\backslash\ldots\backslash X_i\backslash X_{i+1}\backslash
\ldots \backslash S / Y_1 \ldots Y_n \\
\Downarrow \\
X_1\backslash\ldots\backslash X_i\backslash\con{Adj}\backslash X_{i+1}
\backslash
\ldots \backslash S /Y_1 \ldots Y_n
\end{array}
\end{equation}

The derivation of (\ref{adj-ex}a) is given below (where $ X \Rightarrow Y$
indicates that {\em add\_adjuncts}(Y,X) is satisfied, and {\sc iv =
np$\backslash$s}).

{\sc
\begin{equation}
\mbox{
\begin{tabular}[t]{rcccccc}
\ldots {\em J.} & {\em opzettelijk} & {\em een ongeluk} & {\em veroorzaakt}\\
        np         &  adj              &  np  &  np$\backslash$iv \\
                                                &&&$\Downarrow$ \\
                          &&& np$\backslash$(adj$\backslash$iv) \\
\cline{3-4}     && \multicolumn{1}{c}{adj$\backslash$iv} \\
\cline{2-3} & \multicolumn{2}{c}{iv} \\
\cline{1-3}  \multicolumn{2}{c}{s} \\
\end{tabular}
}
\end{equation}
}

An interesting implication of this analysis is that in a categorial setting the
notion `head' can be equated with the notion `main functor'.  This has been
proposed by Barry and Pickering \shortcite{barry-pickering}, but they are
forced
to assign a category containing Kleene-star operators to verbal elements.  The
semantic counterpart of such category-assignments is unclear.  The present
proposal is an alternative for such assignments which avoids introducing new
categorial operators and which does not lead to semantic complications (the
semantics of {\em add\_adjuncts} is presented in section \ref{adj-scope}).
Below we argue that this analysis also allows for a straightforward explanation
of the distribution and scope of adjuncts in verb phrases headed by a verbal
complex.

\section{Cross-Serial Dependencies}

In Dutch, verbs selecting an infinitival complement (e.g.  modals and
perception
verbs) give rise to so called cross-serial dependencies.  The arguments of the
verbs involved appear in the same order as the verbs in the `verb cluster':

\begin{equation}
\label{gramm}
\begin{array}[t]{ll}
a. &
\begin{array}[t]{@{\hspace{0cm}}llllll}
\mbox{dat} & \mbox{An$_1$} & \mbox{Bea$_2$} & \mbox{wil$_1$} &
\mbox{kussen$_2$.}
\\
\mbox{dat} & \mbox{An} & \mbox{Bea} & \mbox{wants} & \mbox{to kiss}
\end{array} \\
& \mbox{\em that An wants to kiss Bea} \\
b. &
\begin{array}[t]{@{\hspace{0cm}}llllll}
\mbox{dat} &  \mbox{An$_1$}&  \mbox{Bea$_2$}& \mbox{Cor$_3$} &\mbox{wil$_1$} \\
\mbox{dat}& \mbox{An}& \mbox{Bea}& \mbox{Cor}& \mbox{wants}
\end{array} \\
& \begin{array}[t]{@{\hspace{0cm}}ll}
\mbox{zien$_2$} &\mbox{kussen$_3$.} \\
\mbox{to see}& \mbox{kiss}
\end{array} \\
& \mbox{\em that An wants to see Bea kiss Cor} \\
\end{array}
\end{equation}

The property of forming cross-serial dependencies is a lexical property of the
matrix verb.  If this verb is a `trigger' for cross-serial word order, this
order is obligatory, whereas if it is not, the infinitival complement will
follow the verb:

\begin{equation}
\label{ungramm}
\begin{array}[t]{ll@{\hspace{0cm}}l}
a. & ^*&\mbox{dat An wil Bea kussen.} \\
b. &&
\begin{array}[t]{@{\hspace{0cm}}lllll}
\mbox{dat}&  \mbox{An}& \mbox{zich}& \mbox{voornam}& \mbox{Bea} \\
\mbox{that}&  \mbox{An}& \mbox{Refl.}& \mbox{planned}& \mbox{Bea}
\end{array} \\
&& \mbox{te  kussen.} \\
&& \mbox{to kiss} \\
&& \mbox{\em that An. planned to kiss Bea} \\
c. &^*&\mbox{dat An zich Bea voornam te  kussen.}
\end{array}
\end{equation}

\subsection{Generalized Division}

Categorial accounts of cross-serial dependencies initially made use of a
syntactic rule of composition \cite{Steedman85}.  Recognizing the lexical
nature
of the process, more recent proposals have used either a lexical rule of
composition \cite{moortgat-diss} or a lexical rule of `division'
\cite{hoeksema-ms}.  Division is a rule which enables a functor to inherit the
arguments of its argument:\footnote{ Argument inheritance is used in {\sc hpsg}
to account for verb clustering in German \cite{HinrichsNakazawa89}.  The {\sc
hpsg} analysis is essentially equivalent to Hoeksema's account.}

\[ X/Y \Rightarrow (X/Z_1 \ldots /Z_n)/(Y/Z \ldots /Z_n)   \]

To generate cross-serial dependencies, a `disharmonic'
version of this rule is needed:

\begin{equation}
\label{gen-div}
X/Y \Rightarrow
(Z_1\backslash\ldots Z_n\backslash X)/(Z_1\backslash\ldots Z_n\backslash Y)
\end{equation}

Hoeksema proposes that verbs which trigger cross-serial word order are subject
to (\ref{gen-div}):

{\sc
\begin{equation}
\mbox{
\begin{tabular}[t]{rccc}
\ldots {\em An} & {\em Bea}     & {\em wil}     & {\em kussen} \\
         np     & np            & iv/iv         & np$\backslash$iv \\
                                && $\Downarrow$ \\
&&  $($np$\backslash$iv$)/($np$\backslash$iv$)$ \\
\cline{3-4}
&& \multicolumn{2}{c}{np$\backslash$iv} \\
\cline{2-3} & \multicolumn{2}{c}{iv}
\end{tabular}
}
\end{equation}
}

In a framework using recursive constraints, generalized disharmonic division
can
be implemented as a recursive constraint connecting the initial category of
such
verbs with a derived category:

\begin{equation}\begin{array}[t]{l}
lex(\con{willen},\var{X}) \con{ :-}\\
\hspace{1cm} \att{cross\_serial}(\var{X},
	(\con{NP}\backslash\con{S})/(\con{NP}\backslash\con{S})).\\
lex(\con{zien},\var{X}) \con{ :-}\\
\hspace{1cm} \att{cross\_serial}(\var{X},
(\con{NP}\backslash(\con{NP}\backslash\con{S}))/(\con{NP}\backslash\con{S})).\\
lex(\con{voornemen},
(\con{NP$_{\mbox{refl}}$}\backslash(\con{NP}\backslash\con{S}))/
						(\con{NP}\backslash\con{S})).
\end{array}
\end{equation}

\begin{equation}\begin{array}[t]{l}
\att{cross\_serial}(\var{Out},\var{In}) \con{ :-}\\
\hspace{1cm} \att{division}(\var{Out},\var{In}),\\
\hspace{1cm} \att{verb\_cluster}(\var{Out}).
\\ \\
\att{division}(X,X). \\
\att{division}((Z\backslash X)/(Z \backslash Y),X^{\prime}/Y^{\prime}) \con{
:-} \\
\hspace{1cm} \att{division}(X/Y,X^{\prime}/Y^{\prime}). \\ \\
\att{verb\_cluster}(\avm{\att{arg } \avm{\att{vc } +}}).
\end{array}
\end{equation}

Only verbs that trigger the cross-serial order are subject to the {\em
division}
constraint.  This accounts immediately for the fact that cross-serial orders do
not arise with all verbs selecting infinitival complements.

\subsection{Verb Clusters}

The {\em verb\_cluster} constraint ensures that cross-serial word order is
obligatory for verbs subject to {\em cross\_serial}.  To rule out the
ungrammatical (\ref{ungramm}a), for instance, we assume that {\em Bea kussen}
is
not a verb cluster.  The verb {\em kussen} by itself, however, is unspecified
for {\sc vc}, and thus (\ref{gramm}a) is not excluded.

We do not assume that cross-serial verbs take lexical arguments (as has
sometimes been suggested), as that would rule out the possibility of complex
constituents to the right of cross-serial verbs altogether.  If one assumes
that
a possible bracketing of the verb cluster in (\ref{gramm}b) is {\em [wil [zien
kussen]]} (coordination and fronting data have been used as arguments that this
is indeed the case), a cross-serial verb must be able to combine with
non-lexical verb clusters.  Furthermore, if a verb selects a particle, the
particle can optionally be included in the verb cluster, and thus can appear
either to the right or to the left of a governing cross-serial verb.  For a
verb
cluster containing two cross-serial verbs, for instance, we have the following
possibilities:

\begin{equation}
\label{prefix}
\begin{array}[t]{ll}
a. &
\begin{array}[t]{@{\hspace{0cm}}lllllll}
\mbox{dat}&  \mbox{An}& \mbox{Bea}& \mbox{heeft}& \mbox{durven} & \mbox{aan}\\
\mbox{that}&  \mbox{An}& \mbox{Bea}& \mbox{has}& \mbox{dared}  & \mbox{part.}
\end{array} \\
& \mbox{te spreken}\\
& \mbox{to speak} \\
& \mbox{\em that An has dared to speak to Bea.} \\
b. & \mbox{dat An Bea heeft aan durven te spreken.} \\
c. &\mbox{dat An Bea aan heeft durven  te spreken.}
\end{array}
\end{equation}

A final piece of evidence for the the fact that cross-serial verbs may take
complex phrases as argument stems from the observation that certain adjectival
and prepositional arguments can also appear as part of the verb cluster:

\begin{equation}
\begin{array}[t]{ll}
\begin{array}[t]{@{\hspace{0cm}}lllllll}
\mbox{dat}&  \mbox{An}& \mbox{dit} & \mbox{aan Bea}& \mbox{had} &
\mbox{duidelijk} \\
\mbox{that}&  \mbox{An}& \mbox{this}& \mbox{to Bea}& \mbox{has}  & \mbox{clear}
\end{array} \\
\mbox{gemaakt}\\
\mbox{made} \\
\mbox{\em that An had made this clear to Bea} \\
\end{array}
\end{equation}

Cross-serial verbs select a {\sc +vc} argument.  Therefore, all phrases that
are
not verb clusters must be marked {\sc -vc}.  In general, in combining a
(verbal)
functor with its argument, it is the argument that determines whether the
resulting phrase is {\sc -vc}.  For instance, {\sc np}-arguments always give
rise to {\sc -vc} phrases, whereas particles and verbal arguments do not give
rise to {\sc -vc} phrases.  This suggests that {\sc np}'s must be marked {\sc
-vc}, that particles and verbs can remain unspecified for this feature, and
that
in the syntactic rule for application the value of the feature {\sc vc} must be
reentrant between argument and resultant.

\subsection{The distribution and scope of adjuncts}
\label{adj-scope}

The analysis of cross-serial dependencies in terms of argument inheritance
interacts with the analysis of adjuncts presented in section \ref{adjuncts}.
If
a matrix verb inherits the arguments of the verb it governs, it should be
possible to find modifiers of the matrix verb between this verb and one of its
inherited arguments.  This prediction is borne out (\ref{interaction}a).
However, we also find structurally similar examples in which the adjunct
modifies the governed verb (\ref{interaction}b).  Finally, there are examples
that are ambiguous between a wide and narrow scope reading
(\ref{interaction}c).
We take it that the latter case is actually what needs to be accounted for,
i.e.
examples such as (\ref{interaction}a) and (\ref{interaction}b) are cases in
which there is a strong preference for a wide and narrow scope reading,
respectively, but we will remain silent about the (semantic) factors
determining
such preferences.

\begin{equation}
\label{interaction}
\begin{array}[t]{ll}
a. &
\begin{array}[t]{@{\hspace{0.0cm}}llllll}
\mbox{dat} & \mbox{Frits} & \mbox{Marie} & \mbox{volgens mij} & \mbox{lijkt} \\
\mbox{that} & \mbox{F.}& \mbox{M.}& \mbox{to me} & \mbox{seems}
\end{array} \\
& \mbox{te ontwijken.} \\
& \mbox{to avoid} \\
& \mbox{\em It seems to me that F. avoids M.} \\
b. &
\begin{array}[t]{@{\hspace{0.0cm}}llllll}
\mbox{dat}& \mbox{Frits} & \mbox{Marie} & \mbox{opzettelijk} & \mbox{lijkt}\\
\mbox{that}& \mbox{F.} &\mbox{M.} & \mbox{deliberately} & \mbox{seems}
\end{array} \\
& \mbox{te ontwijken.} \\
& \mbox{to avoid} \\
& \mbox{\em It seems that F. deliberately avoids M.} \\
c. &
\begin{array}[t]{@{\hspace{0.0cm}}llllll}
\mbox{dat} & \mbox{Frits} & \mbox{Marie} & \mbox{de laatste tijd}
& \mbox{lijkt}\\
\mbox{that}& \mbox{F.} &\mbox{M.} & \mbox{lately} & \mbox{seems} \\
\end{array} \\
& \mbox{te ontwijken.} \\
& \mbox{to avoid} \\
& \mbox{\em It seems lately as if  F. avoids M.} \\
& \mbox{\em It seems as if  F. avoids M. lately}
\end{array}
\end{equation}

On the assumption that the lexical entries for {\em lijken} en {\em ontwijken}
are as in (\ref{lex}), example (\ref{interaction}c) has two possible
derivations
((\ref{wide}) and (\ref{narrow})).  Procedurally speaking, the rule that adds
adjuncts can be applied either to the matrix verb (after division has taken
place) or to the governed verb.  In the latter case, the adjunct is `inherited'
by the matrix verb.  Assuming that adjuncts take scope over the verbs
introducing them, this accounts for the ambiguity observed above.

\begin{equation}
\label{lex}
\begin{array}[t]{l}
lex(\con{lijken},\var{Verb}) \con{:-} \\
\hspace{1cm} \att{add\_adjuncts}(\var{Verb},\var{Verb}^{\prime}), \\
\hspace{1cm} \att{cross\_serial}(\var{Verb}^{\prime},
	(\con{NP}\backslash\con{S})/(\con{NP}\backslash\con{S})).\\
lex(\con{ontwijken},\var{Verb}) \con{:-} \\
\hspace{1cm} \att{add\_adjuncts}(\var{Verb},
	\con{NP}\backslash(\con{NP}\backslash\con{S}) ).
\end{array}
\end{equation}

{\sc
\begin{equation}
\label{wide}
\mbox{
\begin{tabular}[t]{ccc}
\ldots  {\em de laatste tijd} & {\em lijkt}    	& {\em te ontwijken}\\
                adj               & iv/iv  	&  tv \\
                                                &&$\Downarrow$ \\
                                                && tv/tv \\
                                                &&$\Downarrow$ \\
                                && (adj$\backslash$tv)/tv \\
\cline{2-3}     & \multicolumn{1}{r}{adj$\backslash$tv} \\
\cline{1-3}  \multicolumn{2}{c}{tv}
\end{tabular}
}
\end{equation}

\begin{equation}
\label{narrow}
\mbox{
\begin{tabular}[t]{ccccccc}
\multicolumn{3}{c}{\em \ldots de laatste tijd} & {\em lijkt}
						& {\em te ontwijken}\\
      & adj         &          & iv/iv  		&  tv\\
                                        &&&$\Downarrow$ & $\Downarrow$\\
     && \multicolumn{2}{c}{(adj$\backslash$tv)/(adj$\backslash$tv)}
                                & adj$\backslash$tv \\
\cline{3-5}    & & \multicolumn{3}{c}{adj$\backslash$tv} \\
\cline{2-5} &  \multicolumn{3}{c}{tv}
\end{tabular}
}
\end{equation}
}

The assumption that adjuncts scope over the verbs introducing them can be
implemented as follows.  We use a unification-based semantics in the spirit of
Pereira and Shieber \shortcite{pereira-shieber}.  Furthermore, the semantics is
{\em head-driven}, i.e.  the semantics of a complex constituent is reetrant
with
the semantics of its head (i.e.  the functor).  The feature structure for a
transitive verb including semantics (taking two {\sc np}'s of the generalized
quantifier type $\langle\langle e,t \rangle, t\rangle$ as argument and
assigning
wide scope to the subject) is:

\begin{equation}
\label{kussen-sem}
\avm[TV]{	\att{val } \avm{ 	\att{val } \avm{\att{cat } \con{s}}\\
				\att{dir } `\backslash\mbox{'}\\
				\att{arg } \avm{ \att{cat }\con{np}\\
						 \att{sem }
			(\var{X}^{\wedge}\var[Obj]{S})^{\wedge}\var[Subj]{S}
						}
			}\\
	\att{dir } `\backslash\mbox{'}\\
	\att{arg } \avm{	\att{cat } \con{np}\\
			 	\att{sem }
			(\con{Y}^{\wedge}\con{kiss(X,Y)})^{\wedge}\var[Obj]{S}
			}\\
	\att{sem } \var[Subj]{S}
     }
\end{equation}

Thus, a lexical entry for a transitive verb can be defined as follows (where
{\em TV} refers to the feature structure in \ref{kussen-sem}):

\begin{equation}
\begin{array}[t]{l}
lex(\con{kussen},X)  \con{ :-}  \\
\hspace{1cm} \att{add\_adjuncts}(\var{X},\var{TV}).
\end{array}
\end{equation}

The lexical rule for adding adjuncts can now be extended with a semantics:

\begin{equation}
\begin{array}[t]{l}
\att{add\_adjuncts}(\avm[X]{\att{sem } \var[X]{S}},\avm[Y]{\att{sem }
\var[Y]{S}})
\con{ :-} \\
\hspace{1cm} \att{add\_adj}(\var{X},\var{Y},\var[X]{S},\var[Y]{S}). \\
\\
\att{add\_adj}(\con{S},\con{S},\var{Sem},\var{Sem}). \\ \\
\att{add\_adj}(\avm{	\att{val } \var{X} \\
				\att{dir } `\backslash\mbox{'} \\
				\att{arg } \avm{ \att{cat } \con{adj} \\
						 \att{sem }

			{\var[Y]{S}}^{\wedge}\var[A]{S}}
			},
		   \var{Y},
		   \var[X]{S},
		   \var[Y]{S}) \con{ :-} \\
\hspace{1cm} \att{add\_adj}(\var{X}, \var{Y},\var[X]{S},\var[A]{S}). \\ \\
\att{add\_adj}(\avm{	\att{val } \var{X} \\
				\att{dir } \var{D} \\
				\att{arg } \var{A}},
		   \avm{	\att{val } \var{Y} \\
				\att{dir } \var{D} \\
				\att{arg } \var{A}},
		   \var[X]{S},
		   \var[Y]{S}) \con{ :-} \\
\hspace{1cm} \att{add\_adj}(\var{X}, \var{Y},\var[X]{S},\var[Y]{S}). \\
\end{array}
\end{equation}

Each time an adjunct is added to the subcategorization frame of a verb, the
semantics of the adjunct is `applied' to the semantics as it has been built up
so far ($\var[Y]{S}$), and the result ($\var[A]{S}$) is passed on.  The final
step in the recursion unifies the semantics that is constructed in this way
with
the semantics of the `output' category.  As an adjunct $A_1$ that appears to
the
left of an adjunct $A_2$ in the string will be added to the subcategorization
frame of the governing verb after $A_2$ is added, this orders the (sentential)
scope of adjuncts according to left-to-right word order.  Furthermore, since
the
scope of adjuncts is now part of a verb's lexical semantics, any functor taking
such a verb as argument (e.g.  verbs selecting for an infinitival complement)
will have the semantics of these adjuncts in its scope.

Note that the alternative treatments of adjuncts mentioned in section
\ref{adjuncts} cannot account for the distribution or scope of adjuncts in
cross-serial dependency constructions.  Multiple (i.e.  a finite number of)
categorizations cannot account for all possible word orders, since division
implies that a trigger for cross-serial word order may have any number of
arguments, and thus, that the number of `subcategorization frames' for such
verbs is not fixed.  The polymorphic solution (assigning adjuncts the category
{\sc x}$/${\sc x}) does account for word order, but cannot account for narrow
scope readings, as the adjunct will always modify the whole verb cluster (i.e
the matrix verb) and cannot be made to modify an embedded verb only.

\section{Processing}

The introduction of recursive lexical rules has repercussions for processing as
they lead to an infinite number of lexical categories for a given lexical item
or, if one considers lexical rules as unary syntactic rules, to non-branching
derivations of unbounded length.  In both cases, a parser may not terminate.
One of the main advantages of modeling lexical rules by means of constraints is
that it suggests a solution for this problem.  A control strategy which delays
the evaluation of constraints until certain crucial bits of information are
filled in avoids non-termination and in practice leads to grammars in which all
constraints are fully evaluated at the end of the parse-process.

Consider a grammar in which the only recursive constraint is {\em
add\_adjuncts}, as defined in section \ref{adjuncts}.  The introduction of
recursive constraints in itself does not solve the non-termination problem.  If
all solutions for $\att{add\_adjuncts}$ are simply enumerated during lexical
look-up an infinite number of categories for any given verb will result.

During processing, however, it is not necessarily the case that we need to
consider all solutions.  Syntactic processing can lead to a (partial)
instantiation of the arguments of a constraint.  If the right pieces of
information are instantiated, the constraint will only have a finite number of
solutions.

Consider, for instance, a parse for the following string.

{\sc
\begin{equation}
\mbox{
\begin{tabular}[t]{rcccccc}
\ldots {\em J.} & {\em opzettelijk} & {\em een ongeluk} & {\em veroorzaakt}\\
        np         & adj               & np  &  {\em Verb} \\
                                                &&&$\approx$ \\
                          &&& np$\backslash$(adj$\backslash$iv) \\
\cline{3-4}     && \multicolumn{1}{r}{adj$\backslash$iv} \\
\cline{2-3} & \multicolumn{2}{c}{np$\backslash$s} \\
\cline{1-3}  \multicolumn{2}{c}{s} \\
\end{tabular}
}
\end{equation}
}

Even if the category of the verb is left completely open initially, there is
only one derivation for this string that reduces to S (remember that the syntax
uses application only).  This derivation provides the information that the
variable {\em Verb} must be a transitive verb selecting one additional adjunct,
and with this information it is easy to check whether the following constraint
is satisfied:

\[ \mbox{\em add\_adjuncts}
	(\mbox{\sc np$\backslash$(adj$\backslash$(np$\backslash$s))},
	 \mbox{\sc np$\backslash$(np$\backslash$s))}.\]
This suggests that recursive constraints should not be evaluated during lexical
look-up, but that their evaluation should be delayed until the arguments are
sufficiently instantiated.

To implement this delayed evaluation strategy, we used the {\tt block} facility
of Sicstus Prolog.  For each recursive constraint, a {\tt block} declaration
defines what the conditions are under which it may be evaluated.  The
definition
of {\em add\_adjuncts} (with semantics omitted for readability), for instance,
now becomes:

\begin{equation}\begin{array}[t]{l}
\att{add\_adjuncts}(\avm[X]{\att{arg } \var{Arg}},\var{Y}) \con{ :-}\\
\hspace{1cm}\att{add\_adjuncts}(\var{X},\var{Y},\var{Arg}).\\ \\
\mbox{\tt :- block }
\att{add\_adjuncts}(?,?,-).\\ \\
\att{add\_adjuncts}(\con{S},\con{S},\_).\\
\att{add\_adjuncts}(\con{Adj}\backslash \var{X}, \var{Y},\_) \con{ :-}\\
\hspace{1cm} \att{add\_adjuncts}(\var{X}, \var{Y}).\\
\att{add\_adjuncts}(\avm{ \att{val } \var{X}\\
			  \att{dir } \var{D}\\
			  \att{arg } \var{A}},
		   \avm{\att{val }\var{Y}\\
		   \att{dir }\var{D}\\
		   \att{arg }\var{A}},\var{A})\con{ :-}\\
\hspace{1cm}\att{add\_adjuncts}(\var{X}, \var{Y}).
\end{array}\end{equation}

We use {\em add\_adjuncts}/2 to extract the information that determines when
{\em add\_adjuncts}/3 is to be evaluated.  The {\tt block} declaration states
that {\em add\_adjuncts}$/3$ may only be evaluated if the third argument (i.e.
the argument of the `output' category) is not a variable.  During lexical
look-up, this argument is uninstantiated, and thus, no evaluation takes place.
As soon as a verb combines with an argument, the argument category of the verb
is instantiated and {\em add\_adjuncts}$/3$ will be evaluated.  Note, however,
that calls to {\em add\_adjuncts}$/3$ are recursive, and thus one evaluation
step may lead to another call to {\em add\_adjuncts}$/3$, which in its turn
will
be blocked until the argument has been instantiated sufficiently.  Thus, the
recursive constraint is evaluated incrementally, with each syntactic
application
step leading to a new evaluation step of the blocked constraint.  The recursion
will stop if an atomic category {\sc s} is found.

Delayed evaluation leads to a processing model in which the evaluation of
lexical constraints and the construction of derivational structure is
completely
intertwined.

\subsection{Other strategies}

The delayed evaluation techniques discussed above can be easily implemented in
parsers which rely on backtracking for their search.  For the grammars that we
have worked with, a simple bottom-up (shift-reduce) parser combined with
delayed
evaluation guarantees termination of the parsing process.

To obtain an efficient parser more complicated search strategies are
required. However, chart-based search techniques are not easily
generalized for grammars which make use of complex constraints.  Even
if the theoretical problems can be solved
\cite{johnson-memoization,doerre-earley}
severe practical problems might surface, if the constraints are as
complex as the ones proposed here.

As an alternative we have implemented chart-based parsers
using the `non-interleaved pruning' strategy (terminology from
\cite{maxwell-kaplan-cl}).
Using this strategy the parser first builds a parse-forest for a
sentence on the basis of the context-free backbone of the grammar. In
a second processing phase parses are recovered on the basis of the
parse forest and the corresponding constraints are applied. This may
be advantageous if the context-free backbone of the grammar is
`informative' enough to filter many unsuccessful partial derivations
that the parser otherwise would have to check.

As clearly a CUG grammar does not contain such an informative
context-free backbone a further step is to use `selective feature
movement' (cf. again \cite{maxwell-kaplan-cl}).  In this approach the
base grammar is compiled into an equivalent modified grammar in which
certain constraints from the base grammar are converted to a more
complex context-free backbone in the modified grammar.

Again, this technique does not easily give good results for grammars
of the type described. It is not clear at all where we should begin
extracting appropriate features for such a modified grammar, because
most information passing is simply too `indirect' to be easily
compiled into a context-free backbone.

We achieved the best results by using a `hand-fabricated' context-free
grammar as the first phase of parsing. This context-free grammar
builds a parse forest that is then used by the `real' grammar to
obtain appropriate representation(s) for the input sentence. This
turned out to reduce parsing times considerably.

Clearly such a strategy raises questions on the relation between this
context-free grammar and the CUG grammar. The context-free grammar is
required to produce a superset of the derivations allowed by the CUG.
Given the problems mentioned above it is difficult to show that this
is indeed the case (if it were easy, then it probably would also be easy to
obtain such a context-free grammar automatically).

The strategy can be described in somewhat more detail as follows.
The context-free phase of processing builds a number of items defining
the parse forest, in a format that can be used by the second
processing phase. Such items are four-tuples
\[
\langle R, P_0, P, D \rangle
\]
where $R$ is a rule name (consistent with the rule names from the
CUG), $P_0,P$ are string positions and $D$ describes the string
positions associated with each daughter of the rule (indicating which
part of the string is covered by that daughter).

Through a head-driven recursive descent the second processing phase
recovers derivations on the basis of these items. Note that the delayed
evaluation technique for complex constraints is essential here.
Alternative solutions are obtained by backtracking. If the first phase
has done a good job in pruning many failing search branches then this
is not too expensive, and we do not have to worry about the
interaction of caching and complex constraints.

\section{Final Remarks}

In sections 2 and 3 we have sketched an analysis of cross-serial dependency
constructions and its interaction with the position and scope of adjuncts.  The
rules given there are actually part of a larger fragment that covers the syntax
of Dutch verb clusters in more detail.  The fragment accounts for cross-serial
dependencies and extraposition constructions (including cases of `partial'
extraposition), {\em infinitivus-pro-participio}, modal and participle
inversion, the position of particles in verb clusters, clitic climbing, partial
{\sc vp}-topicalization, and verb second.  In the larger fragment, additional
recursive constraints are introduced, but the syntax is still restricted to
application only.

The result of Carpenter \shortcite{carpenter-lexical} emphasizes the importance
of lexical rules.  There is a tendency in both {\sc cg} and HPSG to rely more
and more on mechanisms (such as inheritance and lexical rules or recursive
constraints) that operate in the lexicon.  The unrestricted generative capacity
of recursive lexical rules implies that the remaining role of syntax can be
extremely simple.  In the examples above we have stressed this by giving an
account for the syntax of cross-serial dependencies (a construction that is,
given some additional assumptions, not context-free) using application only.
In
general, such an approach seems promising, as it locates the sources of
complexity for a given grammar in one place, namely the lexicon.


\end{document}